\documentclass[letterpaper, 10 pt, conference]{ieeeconf}  % Comment this line out if you need a4paper

\IEEEoverridecommandlockouts                              % This command is only needed if 
\usepackage{graphics} % for pdf, bitmapped graphics files
\usepackage{epsfig} % for postscript graphics files
\usepackage{mathptmx} % assumes new font selection scheme installed
\usepackage{times} % assumes new font selection scheme installed
\usepackage{amsmath} % assumes amsmath package installed
\usepackage{amssymb}  % assumes amsmath package installed

\usepackage{color}
\usepackage{xcolor}
\usepackage{caption}
\usepackage{preambles}

\title{\LARGE \bf
Velocity Regulation of 3D Bipedal Walking Robots with Uncertain Dynamics Through Adaptive Neural Network Controller}
\author{Guillermo A. Castillo$^{1}$, Bowen Weng$^{1}$, Terrence C. Stewart$^{2}$, Wei Zhang$^{3}$, and Ayonga Hereid$^{4}$% <-this % stops a space
\thanks{*This work was supported in part by the National Science Foundation under grant CNS-1552838, the OSU M\&MS Discovery Theme Initiative, and the startup fund of SUSTech. The authors also  gratefully  acknowledge the Telluride Neuromorphic Cognition Engineering Workshop 2019 for the unique environment that enabled the presented work.}% 
\thanks{$^{1}$Electrical and Computer Engineering, Ohio State University, Columbus, OH, USA;  {\tt\footnotesize \{castillomartinez.2, weng.172\}@osu.edu.}}
\thanks{$^{2}$National Research Council of Canada, University of Waterloo Collaboration Centre, Waterloo ON, Canada; {\tt\footnotesize tcstewar@uwaterloo.ca.}}
\thanks{$^{3}$SUSTech Institute of Robotics, Southern University of Science and Technology (SUSTech), China; {\tt\footnotesize zhangw3@sustech.edu.cn.}}
\thanks{$^{4}$Mechanical and Aerospace Engineering, Ohio State University, Columbus, OH, USA. {\tt\footnotesize hereid.1@osu.edu.}}%
}

\usepackage[capitalise]{cleveref}

\usepackage[noadjust]{cite}

\usepackage{graphicx}

\usepackage[caption=false]{subfig}

\begin{document}
\maketitle
\thispagestyle{empty}
\pagestyle{empty}

\begin{abstract}
% This paper presents an adaptive control structure based in online trained neural networks to regulate the average walking speed of 3D bipedal robots while considering the effect of the dynamic uncertainty of the system due to changes in the dynamic properties of the robot or changes in the environment. Existing Hybrid Zero Dynamics-based controllers regulate walking speed of the robot through the implementation of heuristic regulators that modify the trajectories or torques of the robot's joints. However, since these regulators are not adaptive to the uncertain dynamics of the robot and its environment, the variation of the dynamic properties of the robot affects significantly the speed tracking performance of the controller, leading to steady state velocity errors and some times even to the instability of the walking limit cycle. 

This paper presents a neural-network based adaptive feedback control structure to regulate the velocity of 3D bipedal robots under dynamics uncertainties. Existing Hybrid Zero Dynamics (HZD)-based controllers regulate velocity through the implementation of heuristic regulators that do not consider model and environmental uncertainties, which may significantly affect the tracking performance of the controllers. In this paper, we address the uncertainties in the robot dynamics from the perspective of the reduced dimensional representation of virtual constraints and propose the integration of an adaptive neural network-based controller to regulate the robot velocity in the presence of model parameter uncertainties. The proposed approach yields improved tracking performance under dynamics uncertainties. The shallow adaptive neural network used in this paper does not require training a priori and has the potential to be implemented on the real-time robotic controller. A comparative simulation study of a 3D Cassie robot is presented to illustrate the performance of the proposed approach under various scenarios.

% that modify the trajectories or torques of the robot's joints. However, since these regulators are not adaptive to the uncertain dynamics of the robot and its environment, the variation of the dynamic properties of the robot affects significantly the speed tracking performance of the controller, leading to steady state velocity errors and some times even to the instability of the walking limit cycle. 

\end{abstract}

\section{Introduction}

Model-based controllers for 3D walking robots have received considerable attention from the robotics community due to their ability to take full advantage of the natural hybrid dynamics to achieve dynamic locomotion. Existing approaches, however, may fail to stabilize the robot or accurately track the desired behaviors under model uncertainties. Parameters that can significantly affect system dynamics include the torso's mass, torso's center of mass position, to name a few. On a real robot, these situations may occur when mounting additional equipment in the robot, adding a temporal load, or could be the result of wearing out of mechanical parts like joints or linkages due to the continuous usage of the robot. 
% Such uncertain dynamics could compromise the performance of the system including velocity tracking, energy consumption, and naturalistic walking motion. 

Another important topic in dynamic locomotion is the velocity regulation of the walking robots. Accurate and stable velocity tracking is critical in applications of motion planning, object tracking, and human-robot interactions. For instance, an accompanying robot walking next to a person needs to regulate its velocity to maintain a close distance to that person. Different methods have been proposed to address the velocity tracking problem through feedback controllers based on classical control techniques \cite{Westervelt2003Switching, Da2016From2D, Kobayashi2018Virtual, Chevallereau2003RABBIT, Gasparri2018Efficient}, and more recently based on various machine learning methods \cite{Da2019Combining, Castillo2019Hybrid}. However, these controllers do not take the uncertainties in the robot dynamics into consideration, resulting in noticeably compromised tracking performance in the presence of changes in model properties. For example, in \cite{Chevallereau2003RABBIT}, the authors showed that changing mass and inertia parameters by $\pm 20\%$ results in significant velocity tracking errors.

\begin{figure}
\centering
\vspace{2mm}
\includegraphics[trim={0cm 2.2cm 0cm 2.2cm},clip,width=0.9\columnwidth]{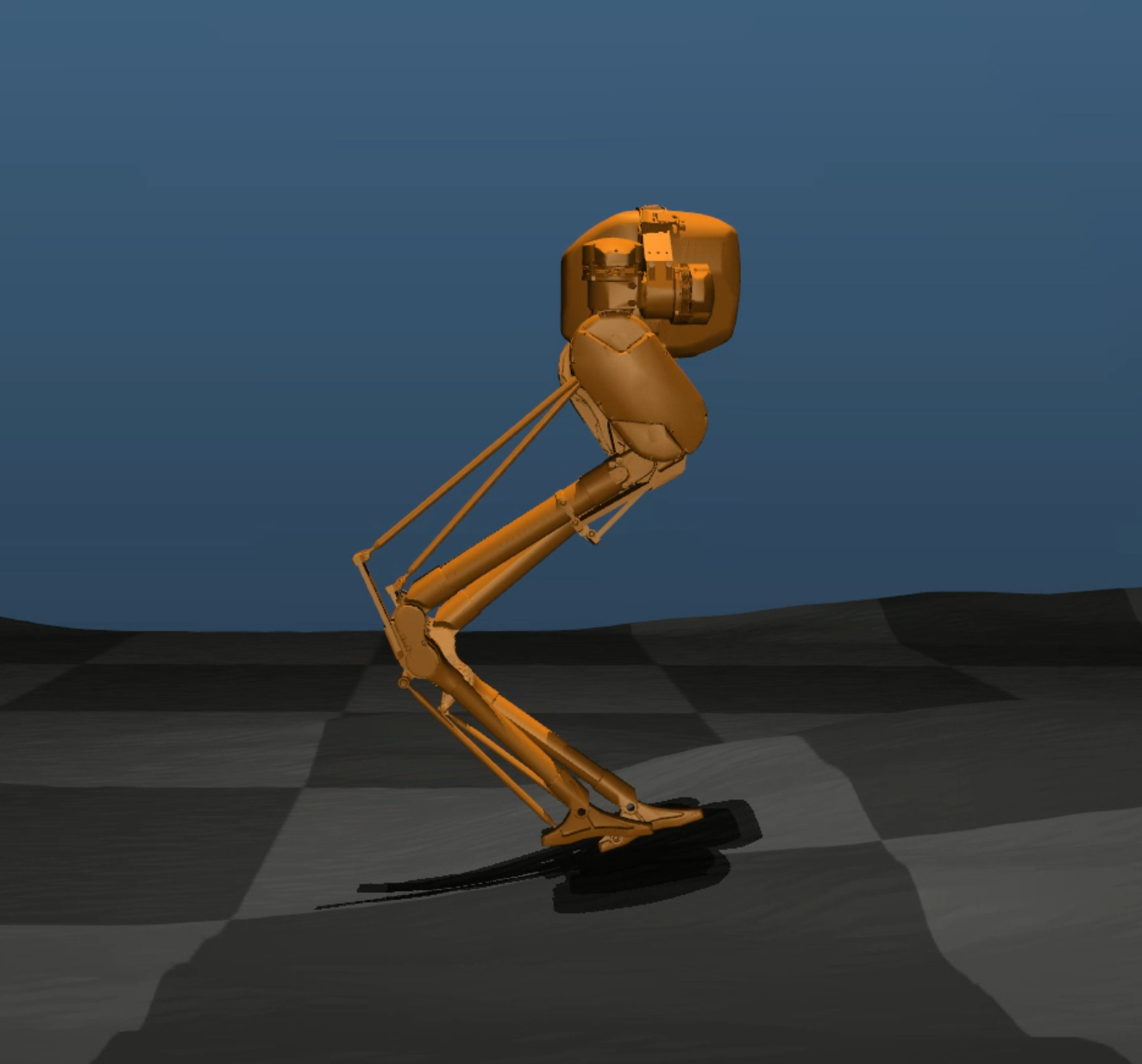}
% \vspace{-7mm}
\caption{Cassie walking on uneven terrain in simulation.} 
\vspace{-3mm}
\end{figure}

% However, neither of these studies have addressed the speed tracking problem when 

% However, when there exists uncertainty in the dynamic properties of the robot or the environment, the walking motion trajectories may not be stable, even in the presence of regulators. In addition, such unknown dynamics would also compromise the performance of the system including speed tracking, energy consumption, and naturalistic walking motion. 
  %The same was true for changes in the impact model of the robot.

% These effects could become particularly noticeable when the variation in the dynamic properties of the robot such as the torso's mass, position of center of mass, or joint damping is significant. 

To understand and simplify the effect of model uncertainties for the stability and performance of bipedal locomotion, researchers have developed bio-inspired heuristic regulators to enhance the rate of convergence to the desired stable limit cycle~\cite{Aoustin2003Control}. In particular, \cite{Aoustin2003Control} showed that the convergence of the nominal reference gait to the cyclic gait regime in planar bipedal robots could be accelerated via feedback regulators on top of the nominal trajectory tracking controllers by varying the torso inclination and the step length. This method is in line with the biomechanical approach for keeping stability while walking: when a person is walking and experiences a disturbance, it is a natural reaction to compensate by simply bending the torso or moving their leg~\cite{Manoonpong2007Adaptive, Aoi2017Adaptive}. 

These findings have been extended to 3D bipedal robots by decoupling the motion of the robot into the longitudinal and lateral planes and applying the foot placement regulator---which varies the step length in response to the robot walking velocity---to track velocity in both directions~\cite{Rezazadeh2015Spring, Da2016From2D}. These heuristically designed regulators often require intensive manual tuning of feedback gains and yield noticeable steady-state tracking errors of velocity when there is a significant change in model parameters.

% To overcome the problems that model-based controllers have for adapting to any changes and uncertainties in the robot's model and the environment, many robot adaptive controllers were proposed during the 80's and 90's \cite{Slotine1987Adaptive, Slotine1989CompositeAC, Ortega1989Adaptive, Woong1997Adaptive}. These adaptive controllers rely on the linear parametrization of the robot dynamics by means of approximating the unknown dynamics of the dynamical system as the linear combination of unknown parameters and a set of basis functions. Then, a control feedback law is computed using the approximation of the unknown dynamics based on the estimation of the unknown parameters, which are updated using an analytical update rule. However, these controllers are restricted to the prior knowledge of the dynamic model and therefore their application is limited when the dynamics of the system is complex and nonlinear. 

Recently, several results have addressed the problem of dynamics uncertainty in bipedal robots using adaptive control from different points of view. In \cite{Gnucci2019OnTheAdaptive}, the authors present a strategy for the non-collocated adaptive control of underactuated mechanical systems by introducing the concept of virtual control and adaptive virtual constraints to produce stable limit walking cycles for a two-link and three-link robots. However, these results are applied only to planar robots, and its extension to 3D complex robots could be limited by the high complexity of their mathematical models. 
%The authors in \cite{Kobayashi2018Virtual}, proposed a method to ensure asymptotic stability of the walking gait through an adaptive speed controller. This approach uses the inverted pendulum as a reduced order representation to design a virtual dynamics, which can be adapted according to a falling risk of the robot. 

In this work, we are specifically interested in addressing uncertainty in the robot dynamics and its effect on the robot's walking speeds. Therefore, we attempt to provide a general framework to stabilize the robot's walking limit cycle while tracking a desired average walking speed by employing adaptive neural network-based controllers. Inspired by the effective results of neural network controllers in the adaptive control of robot manipulators, we propose a control structure that combines the advantages of online learning with the robustness of classical feedback controllers to compensate for changes in the dynamic properties of the robot. It is worth mentioning that the controller proposed in this work is conceived as an adaptation technique to modify limit walking cycle gait trajectories already obtained by existing trajectory planning algorithms. Therefore, we do not focus on the process of obtaining such trajectories. Instead, we present a novel structure of adaptive feedback regulators, which augment the nominal gaits trajectories rendering stable walking limit cycles at any desired walking speed within a wide range despite the changes in model parameters. 

We further summarize the primary contributions of the present paper as follows:
\begin{itemize}
    %\item  \textbf{Model-free.} A novel model-free method to realize stable and robust 3D bipedal walking using a hybrid learning-control approach that combines the advantages of RL algorithms and classic feedback controllers.
    \item A novel velocity tracking controller for bipedal walking is proposed. Through a nonlinear neural network parameterization, the proposed adaptation scheme is implemented online along with the nominal controller with fast convergence to steady-state velocity under various dynamic uncertainties. In addition, note that the proposed adaptation framework is compatible with nominal controllers derived in both model-free and model-based fashions, and its simple structure makes it feasible to implement on real-time controllers.
    \item The proposed method yields consistent performance against significant uncertainties of dynamic properties. In particular, we show the robust performance of the adaptive controller with pelvis mass experiencing up to 130\% of the mass increase, and the center of mass assigned with up to $0.1$ m offset.%, and the joint damping change by a factor of 10.
    \item The adaptation scheme does not require any dynamic or kinematic properties of the robot. The prorogation of the adaptive controller only relies on observable states and measurable tracking error.
\end{itemize}

% Though we present simulation results, different to many other RL methods proposed to bipedal walking, we show our method is applicable to real 3D robots with highly dynamic motions, such as Cassie. We backup this statement by providing results of the control input for each actuator, showing they are smooth signals that can be applied to real robots. 

The remainder of the paper is organized as follows. \secref{section2} reviews the basics of hybrid zero dynamics and heuristic regulators for velocity regulation. \secref{section3} present our main result of the paper, a novel neural-network based adaptive feedback regulators for dynamics uncertainties, and \secref{section4} shows the improved performance of the proposed approach on 3D Cassie robot in simulation. Conclusions are given in \secref{sec:conclusion}.

\section{PROBLEM FORMULATION} \label{section2}

In this section, we first describe the classical structure of the HZD-based feedback controllers for 3D walking robots. Then we discuss how heuristic regulators can be designed to track velocity under model uncertainties.

\subsection{HZD-based Feedback Controllers}
\label{sec:hzd-control}
The HZD framework provides conditions for the existence of provably stable limit walking cycles by enforcing virtual constraints that are invariant through impact.
%using parameter offline optimization with the full-order model of the robot. 
This technique allows synthesizing feedback controllers that realize stable and dynamic locomotion in 2D/3D robots~\cite{Westervelt2007Feedback, Ames2014Human}.

% \newsec{Robot Model:}
% The Cassie-series bipedal robot, designed by Agility Robotics, is an underactuated biped with 20 degrees of freedom (DOF) in total and highly dynamic capabilities. Each leg has seven joints, in which electric motors directly actuate five of them, and the other two joints are connected via specially designed leaf-spring four-bar linkages for additional compliance. 

\newsec{Virtual Constraints:} Let $\mathbf{q}$ be the vector of the joint coordinates of the robot. The virtual constraints are defined as the difference between the actual and desired outputs of the robot~\cite{Ames2014Human}:
\begin{align}
  \vspace{-5mm}
  \mathbf{y}_{2} &:= \mathbf{y}^a_{2}(\mathbf{q}) - \mathbf{y}^d_{2}(\tau,\alpha),
  \label{eq_vc}
\end{align}
where $\mathbf{y}^d_{2}$ is given as a vector of B\'ezier polynomials parameterized by the coefficients $\mathbf{\alpha}$, and $\tau \in [0,1] $ is the phase variable that synchronizes all virtual constraints.
%, given as:
% \begin{align}
%     \mathbf{y}^d_2(\tau(t),\mathbf{\alpha}) := \sum_{k=0}^{5} \alpha[k] \frac{M!}{k!(M-k)!} \tau(t)^k (1-\tau(t))^{M-k}.
% \end{align}
In this paper, we choose $\tau$ to be the scaled relative time during one walking step, i.e.,
\begin{align}
    \label{eq:tau}
    \vspace{-5mm}
    \tau(t) = \frac{t - t^-}{t_{step}},
\end{align}
where $t_{step}$ is the duration of one walking step, and $t^-$  is the time at the beginning of the step. Typically, the coefficients of desired outputs are determined via model-based offline gait optimization to achieve different periodic walking motions~\cite{Hereid2018Dynamic}.

\subsection{Dynamics Uncertainties and Heuristic Regulators} 
\label{sec:effect}
% In general, HZD-based controllers realizes provably stable periodic walking through enforcing virtual constraints that are obtained via model-based gait optimization.
% The complex dynamics of 3D bipedal robots, however, presents a substantial  challenge when designing controllers using the full-order model of the robot. 
The mismatch between the mathematical model and the real model of 3D bipedal robots leads to the failure of the designed controllers when applied to the real hardware. To prevent from falling, researchers proposed to use heuristically-designed feedback regulators on top of the HZD-based designed controllers to realize asymptotic stability of the walking cycle ~\cite{Reher2016Algorithmic,Gong2019Feedback}.
% Nevertheless, we know that should we be able to consider all the unknown dynamics in our mathematical model, we could find an stable orbit that works for the real robot. Then, the use of additional regulators on top of the HZD-based designed controllers is indeed necessary to find these "new" stable trajectories. 

In practice, two regulators are mostly used to stabilize the walking motion of a 3D bipedal robot: foot placement and torso regulators~\cite{Da2016From2D, Rezazadeh2015Spring, Gong2019Feedback}. These regulators often use decoupled structures, relating certain joints with a specific desired feature---such as hip velocity or torso inclination---of the robot motion. 
% Here, we will focus in the foot placement regulator as it is the most important to regulate the walking speed of the robot. 
% The reader can refer to ~\cite{Da2016From2D, Rezazadeh2015Spring, Gong2019Feedback} for additional information about the other regulators.
%\newsec{Foot placement} controller has been widely used to improve the speed tracking and the stability and robustness of the walking gait. 
For example, the foot placement regulator adds an offset to the desired swing hip pitch and swing hip roll joints to regulate the longitudinal and lateral walking speeds, respectively, and prevent the robot from falling. These offsets are determined by
\begin{align}
    \label{eq:long_reg}
    \hspace{-3em}
    \delta^{sw}_{hpitch}[k] &= K_{p_{x}}(v_{x}[k]-v^{d}_{x}) + K_{d_{x}}(v_{x}[k]-v_{x}[k-1]),\\
    \label{eq:lat_reg}
    \delta^{sw}_{hroll}[k] &= K_{p_{y}}(v_{y}[k]-v^{d}_{y}) + K_{d_{y}}(v_{y}[k]-v_{y}[k-1]),
\end{align}
where $v_{x}[k]$ and $v_{y}[k]$ are the average longitudinal and lateral speeds of the robot at the middle of step $k$, $v^{d}_{x}$, $v^{d}_{y}$ are the reference speeds, and $K_{p_{x}}, K_{d_{x}}, K_{p_{y}}, K_{d_{y}}$ are manually tuned gains. The readers can refer to ~\cite{Da2016From2D, Rezazadeh2015Spring, Gong2019Feedback} for additional information about the other regulators.

\section{Approach} \label{section3}

In this section, we propose a non-conventional controller structure using an adaptive neural network to realize stable walking while tracking desired walking speeds in the presence of changes in the model properties. 

\subsection{Motivation}
\label{sec:integrating_NN}
%From a biomechanics perspective, several studies have shown that the ability of humans to adapt quickly to terrain changes and to walk differently under particular circumstances arise from a combination of biomechanics and neuronal control \cite{Manoonpong2007Adaptive, Aoi2017Adaptive}. That is, motor learning is continuously occurring in the human system to adapt to different walking behaviors or environments. Therefore, it is natural to think that online learning methods could be applied to enhance the abilities of walking robots to adapt to unknown internal dynamics and environments.

%In this work, we intend to address the effect of the unknown dynamics of the robot in the speed tracking performance of HZD-based controllers. 
Motivated by the work of \cite{Narendra1997Adaptive, Lewis1995Neural, Sanne1995Stable, Chen1992Adaptive} in the application of neural networks for the control of nonlinear systems e.g., robotics manipulators, we propose a  framework of adaptive neural network-based controllers that compensate the unknown dynamics of the system while tracking the desired velocity. In addition, we show that the proposed network could be seen as a generalization of the heuristic regulators described in \secref{sec:effect}, and its extension to the case of unknown dynamics. By this, we aim to develop a new general framework for the development of adaptive feedback controllers that render stable limit walking cycles, even when the dynamic properties of the robot change. To validate the proposed method in simulation, we use as our testbed the Cassie-series bipedal robot described in \secref{sec:hzd-control}. 

\subsection{A Review of Adaptive Neural Network-based Controller}
    % The kinematics and dynamics of modern robotic systems are highly complex and nonlinear. The assumption of having exact models 
Typical adaptive controllers rely on approximating the unknown dynamics of the dynamical system as the linear combination of unknown parameters of the system. Then, a control feedback law is computed using the approximation of the unknown dynamics based on the estimation of the unknown parameters, which are updated using a close form update rule. In particular, \cite{Slotine1987Adaptive} showed that the unknown dynamics of a robotic manipulator could be approximated as 
\begin{align}
    \label{eq:approx_din}
    % \hspace{-3em}
    \tilde{\tau} = \mathbf{Y}(\mathbf{q}) \mathbf{\omega},
\end{align}
where $\tilde{\tau}$ is the approximated unknown dynamics, $\mathbf{Y}(\mathbf{q})$ is a matrix of functions dependent on the state of the robot, and $\mathbf{\omega}$ is a vector of parameters. Then, an update law of the form 
\begin{align}
    \label{eq:update_law}
    % \hspace{-3em}
    \dot{\hat{\omega}} = - \mathbf{K} \mathbf{Y}^\intercal \mathbf{s},
\end{align}
is used to estimate the unknown parameters online, where $\mathbf{K}$ is a symmetric positive definite matrix, and $\mathbf{s}$ is the filtered tracking error. However, these controllers are restricted to the prior knowledge of the dynamics structure and extensive system modeling and preliminary analysis are required to compute the regression matrix. 
% Therefore the following adaptive control law was proposed for the control of these type of robotic systems:
% \begin{align}
%   u = K_p*(q_d-q) + K_d*(\dot{q}_d - \dot{q}) + u_{adap}
% \end{align}

% With the emergence of learning methods and the ability of these methods to construct complex, nonlinear function approximators, many new control algorithms using online and offline learning were proposed \cite{Chen1992Adaptive, Lewis1995Neural, Sanne1995Stable}. These methods proposed to use neural networks as function approximators for the unknown dynamics of the system, and proved that for a specific type of systems, adaptive controllers using neural networks can guarantee the convergence of the tracking error to zero, while guaranteeing the boundedness of the error and weights of the neural network. For this, particular update laws for the weights of the neural network were proposed . 

With the emergence of learning methods and the ability of neural networks to approximate complex, nonlinear functions, new neural network-based adaptive control algorithms were proposed  \cite{Chen1992Adaptive, Lewis1995Neural, Sanne1995Stable, Narendra1997Adaptive}. The main advantage of neural network-based controllers is that they can virtually approximate any smooth functions, including the unknown dynamics in a robotic system, without the need for computing a regression matrix. Analogous to equation \eqref{eq:approx_din}, the unknown dynamics of more complex systems, for which a linear parameterization is not accurate enough, can be effectively approximated using neural networks. Then, the nonlinear parameterization of the unknown dynamics can be obtained through neural networks as
\begin{align}
    \label{eq:nn_approx}
    % \hspace{-3em}
    \hat{f}(\mathbf{x}) = \hat{\mathbf{W}}^\intercal \sigma (\hat{\mathbf{V}}^\intercal \mathbf{x}),
\end{align}
where $\hat{{\mathbf{W}}}$, $\hat{{\mathbf{V}}}$ are estimates of the ideal neural network weights provided by some on-line weight tuning algorithms.
% One of the most used tuning algorithms in on-line learning is the delta rule, which will be described in \secref{section3}.

\subsection{Adaptive Feedback Regulators for Virtual Constraints}

% The authors in \cite{Aoustin2003Control}, show that the regulation of the step length and torso inclination helps to accelerate the convergence of the walking limit cycle and that feedback regulators can be designed achieve such convergence. This a very important result that has been the foundation for the use of additional regulators in the HZD. Thus, based in these results, and their extension on \cite{Da2016From2D, Rezazadeh2015Spring, Gong2019Feedback} for 3D walking robots, we decouple our analysis of the virtual constraints into 3 main groups: step length on the longitudinal plane, step length on the sagital plane, and torso inclination. 

% focus our analysis in the re
% decouple the speed tracking problem into the longitudinal and sagital plane.

\begin{figure}
\centering
\vspace{2mm}
\includegraphics[trim={0cm 0cm 0cm 0cm},clip,width=1\columnwidth]{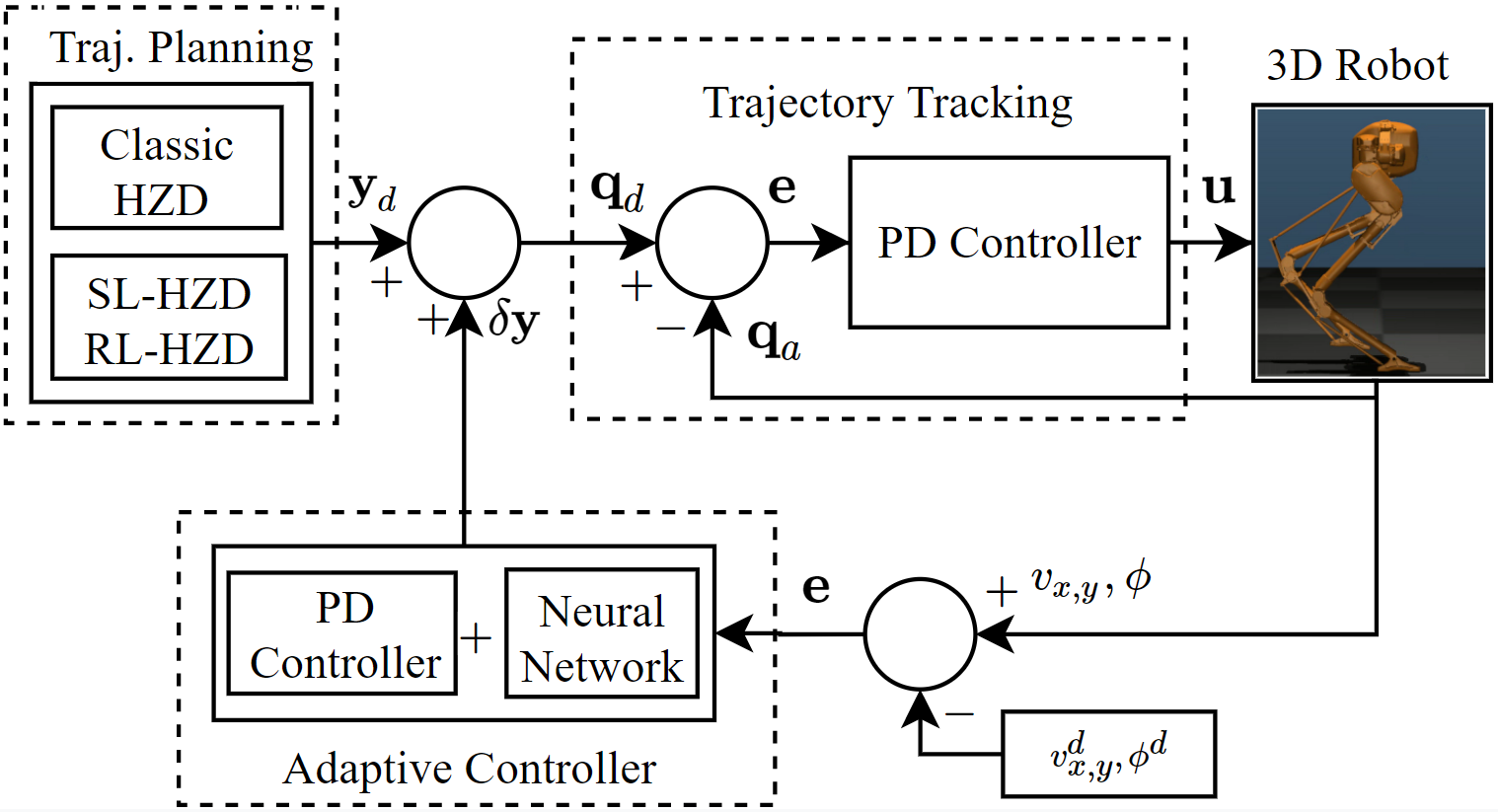}
% \vspace{-7mm}
\caption{The feedback structure of the adaptive regulator integrated into the HZD control framework.} 
\label{structure}
\vspace{-3mm}
\end{figure}

% .  we focus our attention in the regulation of the trajectories that are directly related with the step length.
In this section, we propose an adaptive feedback regulator for virtual constraints to achieve improved speed tracking under model uncertainties. The proposed regulator will have the following form:
\begin{align}
  \mathbf{y}_{2} &:= \mathbf{y}^a_{2}(\mathbf{q}) - (\mathbf{y}^d_{2}(\tau(t),\mathbf{\alpha}) + \delta \mathbf{y}^d_{2}),
  \label{eq_vc_deltay}
\end{align}
where $\delta \mathbf{y}^d_{2}$ represent the modification in the original trajectory to render a stable walking limit cycle, which can be obtained either from offline optimization~\cite{Ames2014Human,Gong2019Feedback}, or offline training of a neural network policy~\cite{Castillo2019Hybrid,Da2019Combining}.

\begin{figure}
\centering
\vspace{2mm}
\includegraphics[trim={0cm 0cm 0cm 0cm},clip,width=0.9\columnwidth]{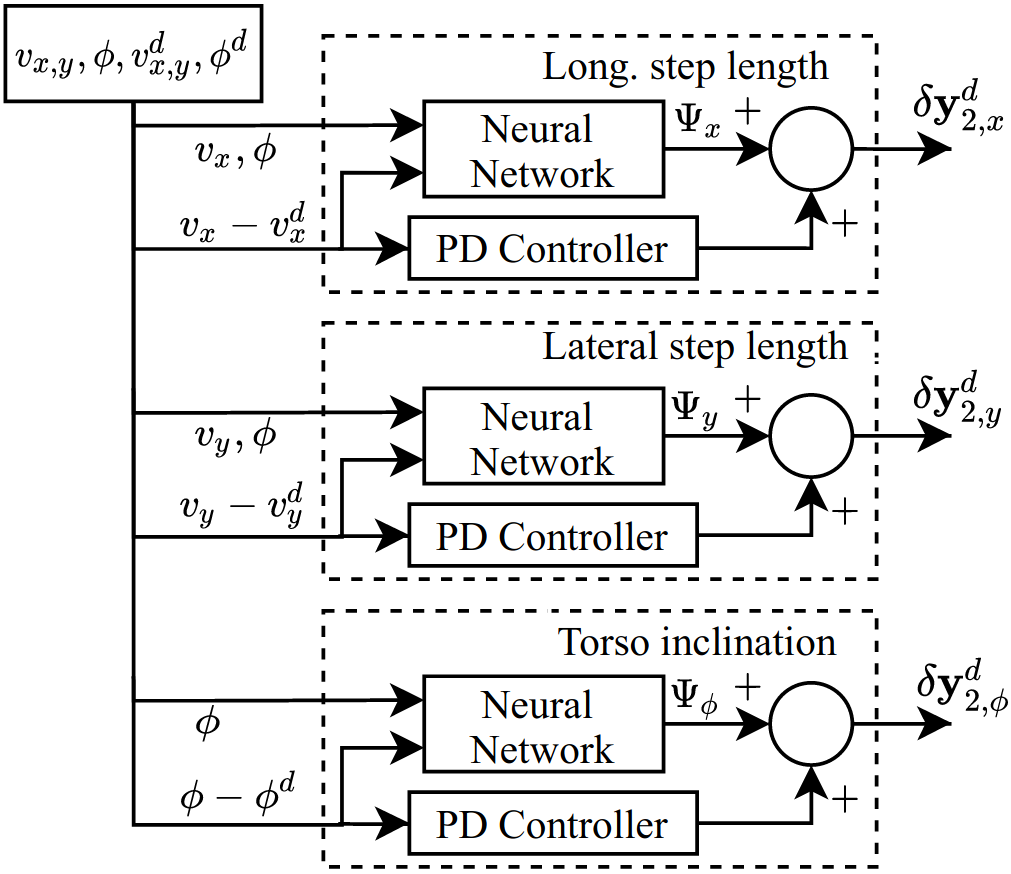}
% \vspace{-7mm}
\caption{Detailed structure of the proposed adaptive neural network-based controller.} 
\label{struct_adap_control}
\vspace{-3mm}
\end{figure}

We will then use an adaptive neural network based feedback controller to determine $\delta \mathbf{y}^d_{2}$, as shown in \figref{structure}. 
Inspired by the results of \cite{Aoustin2003Control}, we focus our analysis on three specific outputs: i) the virtual constraints related to the robot's joints that control the step length in the longitudinal plane (swing hip pitch angle, swing knee), ii) the virtual constraints related to the joints that control the step length in the frontal plane (swing hip roll angle, stance hip roll angle), and iii) the virtual constraints related with the joints that control torso inclination (stance hip pitch angle, stance knee, stance hip roll angle). Let $v_{x}[k]$ and $v_{y}[k]$ be the average longitudinal and lateral speeds of the robot in the middle of step $k$, $v^{d}_{x}$, $v^{d}_{x}$ be the reference speeds, $\phi,\phi^d, \dot{\phi}, \dot{\phi}^d$ be the actual and desired torso inclination and angular velocity, we define
\begin{equation}
    \begin{aligned}
        \label{eq_deltay_est}
        {\delta \mathbf{y}}^d_{2,x} &= K_{p_{x}}(v_{x}[k]-v^{d}_{x}) + K_{d_{x}}(v_{x}[k]-v_{x}[k-1]) + \Psi_{x}, \\
        {\delta \mathbf{y}}^d_{2,y} &= K_{p_{y}}(v_{y}[k]-v^{d}_{y}) + K_{d_{y}}(v_{y}[k]-v_{y}[k-1]) + \Psi_{y}, \\
        {\delta \mathbf{y}}^d_{2,\phi} &= K_{p_{\phi}}(\phi - \phi^{d}) + K_{d_{\phi}}(\dot{\phi} - \dot{\phi}^{d}) + \Psi_{\phi}.
    \end{aligned}
\end{equation}    
where ${\delta \mathbf{y}}^d_{2,x}$, ${\delta \mathbf{y}}^d_{2,y}$, ${\delta \mathbf{y}}^d_{2,\phi}$ represent the modification that compensates the unknown dynamics corresponding to each of the three decoupled systems respectively. 
A detailed structure for the decoupled controllers is presented in \figref{struct_adap_control}, each of which resembles a feedback PD controller and a feed-forward neural network term to track desired behaviors under uncertainties. This adaptive controller structure enhances the robustness of the controller since it allows the PD term to keep the system stable while the network is learning to compensate changes in the dynamic properties of the robot or the environment. The details of the structure and update rule of the neural networks that compute $\Psi_{x}, \Psi_{y}, \Psi_{\phi}$ will be discussed in the following section.

% The structure of the controller resembles a PD controller with a feed-forward term, where $v_{x}[k]$ and $v_{y}[k]$ are the average longitudinal and lateral speeds of the robot at the middle of step $k$, $v^{d}_{x}$, $v^{d}_{x}$ are the reference speeds, $\phi,\phi^d, \dot{\phi}, \dot{\phi}^d$ are the actual and desired torso inclination and angular velocity, $K_{p}, K_{d}$ are manually tuned gains, and $\Psi_{x}, \Psi_{y}, \Psi_{\phi}$ are the outputs of the neural networks used to compensate the effect of the unknown dynamics in the system.

Finally, we denote that the structure chosen for the adaptive controllers proposed in \eqref{eq_deltay_est} allows us to generalize the use of the additional regulators to stabilize the walking limit cycle described in \secref{section2}.
% , and formalize them within the proposed framework of adaptive control.  
In particular, when the output of the neural network is zero, the adaptive controller renders the structure of the traditional regulators. However, the generalized structure proposed by this adaptive controller does not restrict the regulation of the joint trajectories to only the swing hip pitch angle (as in the traditional approach) but allows the controller to learn which trajectories should be modified in order to achieve the successful regulation of the desired longitudinal and lateral velocity. \secref{section4} illustrates this point in detail through simulation results on the bipedal robot Cassie under various scenarios. 
% Maybe include torso here as well and reinforce the connection between them and the accelerated convergence of the walking limit cycle.

\subsection{Adaptive Neural Network Structure} 

% The structure of the Neural Network used in the adaptive controller is inspired by the update rule proposed in \cite{Slotine1987Adaptive}, and has a very simple structure, which makes the proposed method easy to implement and efficient. 

% The structure of the neural network is presented in \figref{neural_network}. 
As shown in \figref{struct_adap_control}, the inputs of the neural networks are the actual and desired values of the longitudinal and lateral velocity, torso inclination, and torso angular velocity. The outputs are the feedforward terms compensating for the unknown dynamics of the decoupled systems. Each network only has one hidden layer with one thousand neurons.  Notice that we can think of the hidden layer neurons as creating a random set of basis functions, and the task of the neural network is to learn the weights on those basis functions that provide the desired compensation as a function of the inputs to the matrix. This is a variant of the adaptive controller proposed in \cite{Slotine1987Adaptive}, and can be given as a simple delta-rule that only applies to the output weights:
% \begin{figure}
% \centering
% \vspace{1mm}
% \includegraphics[trim={0cm 0cm 0cm 0cm},clip,width=1\columnwidth]{figures/struct_adap_controller.png}
% % \vspace{-7mm}
% \caption{Detailed structure of the proposed adaptive neural network-based controller.} 
% \label{neural_network}
% \vspace{-3mm}
% \end{figure}
\begin{align}
  \Delta w_{i,j} = -\gamma E_j h_i,
  \label{eq_learning_rule}
\end{align} 
where $w_{i,j}$ is the weight from the $i$th hidden neuron to the $j$th output, $E_j$ is the error signal for the $j$th output (the difference between the desired and current velocity), $h_i$ is the output of the $i$th hidden neuron, and $\gamma$ is the learning rate chosen as $\gamma = 1e^{-4}$.  This is structurally the same as~\eqref{eq:update_law}.

It is important to mention that the neural network does not require training a priori since the learning process is performed online. The output weights are initialized to zero, and the input weights are randomly initialized.  We use the encoder initialization scheme from \cite{Eliasmith2003} to generate input weights with a broad distribution, ensuring a collection of basis functions that covers the space.  This general approach has been used as a model of biological adaptive arm control \cite{DeWolf2016} and as a simple benchmark task controlling an inverted pendulum \cite{Stewart2015}, but here we apply it in a very different context and with a different source of the error signal.  Previous applications had always been in the domain of torque control and used the output of a PD controller to generate the error signal.

\section{SIMULATION RESULTS} \label{section4}
The proposed method is validated in a dynamic simulation of Cassie using Mujoco ~\cite{Todorov2012MuJoCo}. This section presents the results of the adaptive neural network based controller when the robot is subject to changes in the dynamic properties of the robot such as variation in the mass and center of mass (COM) position of the robot's torso. We also present a comparison of the adaptive controller with the traditional HZD controller and HZD-based RL controller. Finally, we test the robustness of the controller when the robot is subject to adversarial forces and walking on uneven terrain. These results of the evaluation of the proposed adaptive controller can be visualized in the supplemental video material accompanying this submission \cite{video_link}.

% We used the model information of Cassie robot provided by Agility Robotics and the Oregon State University, which is publicly available~\cite{AgilitySimsJune2018}. The number of trainable parameters for the NN is 5069, and the training time is about 10 hours using a single 12-core CPU machine. Visualized results of the learning process and evaluation of the policy in simulation can be seen in the accompanying video submission~\cite{video_link}. This section presents the performance of the control policy obtained from the training in terms of (i) speed tracking, (ii) disturbance rejection, and (iii) the convergence of stable periodic limit cycles. 

%In order to encourage a good speed tracking performance on the learned policy, for each episode of rollout, the desired velocity is updated once at the beginning of the episode (uniform sampling). \ayonga{this description should be in the previous section.}

\subsection{Response to changes in the model properties of the robot}
% Due to the decoupled structure, the learned controller can effectively track a wide range of desired walking speeds in both longitudinal and lateral directions. The performance of tracking a fixed desired speed of 0.5 $m/s$ in the forward direction is shown in \figref{fix_vel}. From \figref{fix_vel}(c), one can see the controller is capable of keeping the upright position of the torso while walking. This particular behavior is encouraged by the reward function during the training process, and it also contributes to the stability of the walking gait.

\begin{figure}
\centering
\vspace{1mm}
\includegraphics[trim={0cm 0cm 0cm 0cm},clip,width=0.9\columnwidth]{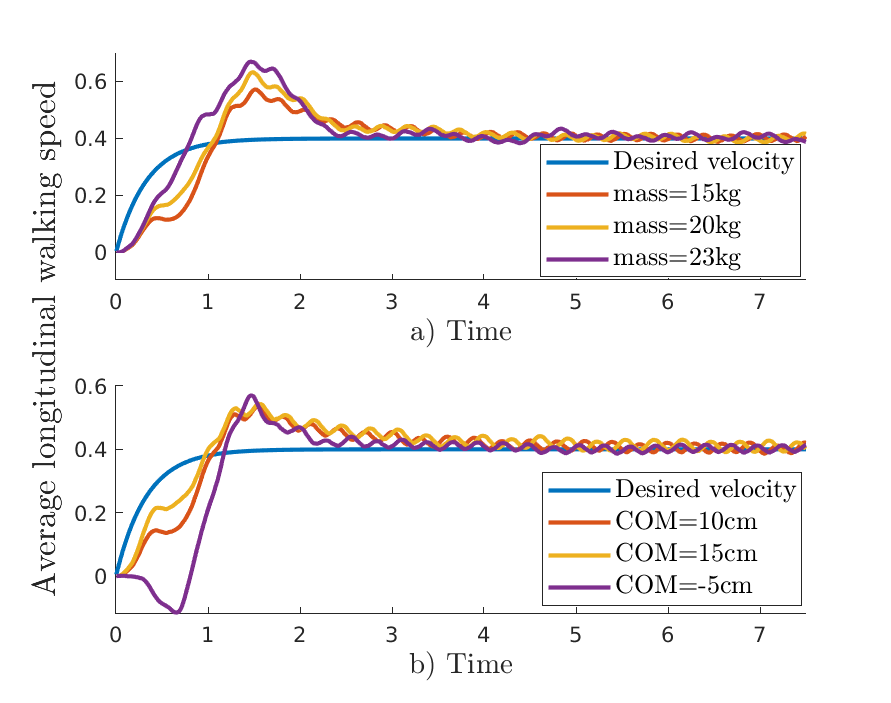}
% \vspace{-7mm}
\caption{The response of of the adaptive controller when (a) the torso's mass is increased; (b) the torso's center of mass position shifts in the longitudinal direction.} 
\label{mass_com_change}
\vspace{-3mm}
\end{figure}

First, we tested the response of the controller when the mass of the torso is increased by approximately $50\%$, $100\%$ and $130\%$. The original mass of the robot's torso is $10.33 \: kg$; then, after the increments in mass the total mass of the torso corresponds to  $15 kg$, $20 kg$, and $23 kg$ respectively. \figref{mass_com_change} (a) shows the response for these three cases. Interestingly, we can see that the tracking error of the average speed does not converge to zero immediately, but it takes some time until the controller learns to compensate for the unknown dynamics, which illustrates the on-line learning process of the adaptive controller.

We also tested the controller with changes in different dynamic properties like the position of the center of mass of the pelvis, by adding an offset of $+0.05 m$, $+0.1 m$,  and $-0.1 m$ in the longitudinal direction. The responses of the adaptive controller for the 3 cases are shown in \figref{mass_com_change} (b), where we can see that the controller performs well even under large parameter uncertainties of the robot's dynamics. Similarly as in the previous test, we can see the learning curve of the controller while the actual walking velocity converges to the desired velocity.

%\begin{figure}
%\centering
%\vspace{1mm}
%\includegraphics[trim={0cm 0cm 0cm %0cm},clip,width=0.9\columnwidth]{figures/com_change.png}
%\caption{Performance of the adaptive controller when the robot's torso COM changes.} 
%\label{COM_change}
%\vspace{-3mm}
%\end{figure}

\subsection{Comparison with traditional and RL HZD-based controller}

\begin{figure}
\centering
\vspace{1mm}
\includegraphics[trim={0cm 0cm 0cm 0cm},clip,width=1\columnwidth]{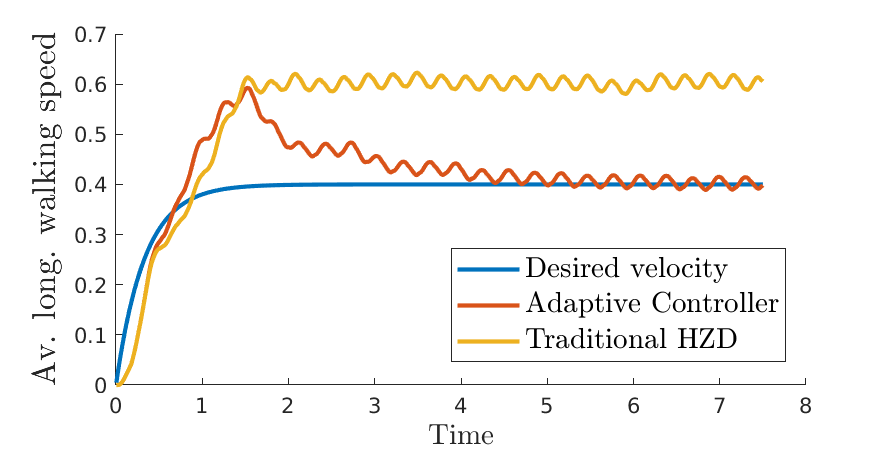}
\caption{Velocity tracking performance of the proposed adaptive controller compared against the classic HZD-based controller. A significant steady-state tracking error can be observed when the adaptive controller is not being used.} 
\label{comparison_HZD}
% \vspace{-3mm}
\end{figure}

\begin{figure}
\centering
\vspace{1mm}
\includegraphics[trim={0cm 0cm 0cm 0cm},clip,width=1\columnwidth]{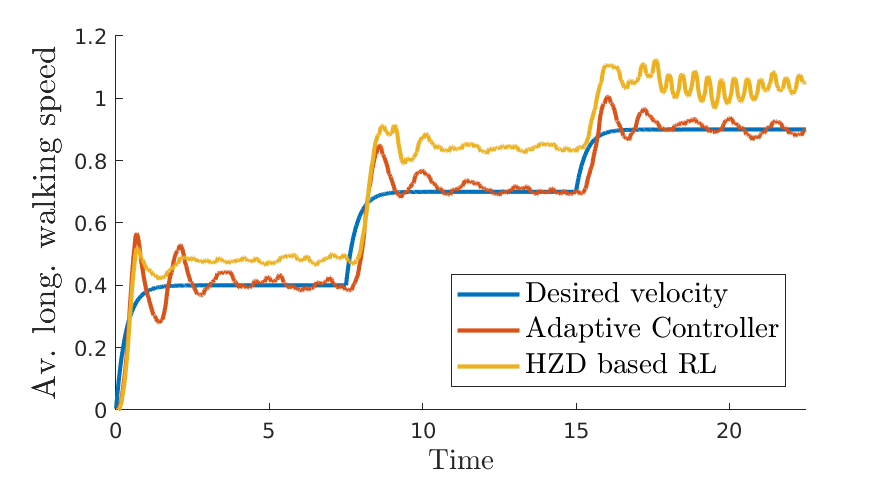}
\caption{The steady-state error of the HZD-based RL controller is more noticeable as the desired walking velocity increases whereas the proposed adaptive controller has almost the same convergence performance regardless of the walking velocity.} 
\label{comparison_RLHZD}
% \vspace{-3mm}
\end{figure}

To illustrate the significant contribution of the adaptive controller to the speed tracking and stability of the system in the face of model uncertainties, we compared its performance against two HZD-based controllers. \figref{comparison_HZD} and \figref{comparison_RLHZD} show the velocity tracking performance of the adaptive controller when compared with the classic HZD-based controller for tracking fixed desired velocity~\cite{Gong2019Feedback}, and with an HZD-based Reinforcement Learning controller for tracking varying velocities 
\cite{Castillo2019Hybrid}. The robot's pelvis mass is increased by $5 kg$, and an offset of $+0.1 m$ is added to the pelvis' COM in the longitudinal direction. As shown in \figref{comparison_HZD} and \figref{comparison_RLHZD}, the existing two HZD-based controllers yield significantly large steady-state tracking error when the dynamics properties of the robot change relative to the model used in the design of these controllers. With the adaptive controller, however, the actual walking velocities converge to a wide range of desired velocities under these changes. More importantly, the convergence is achieved through online learning of the model and does not require a priori training.
% In both cases, it is clear that the adaptive controller outperforms the existing HZD-based controllers, not only in terms of speed tracking but also in keeping the stability of the walking limit cycle. 

\begin{figure}
\centering
\vspace{1mm}
\includegraphics[trim={0cm 0cm 0cm 0cm},clip,width=0.9\columnwidth]{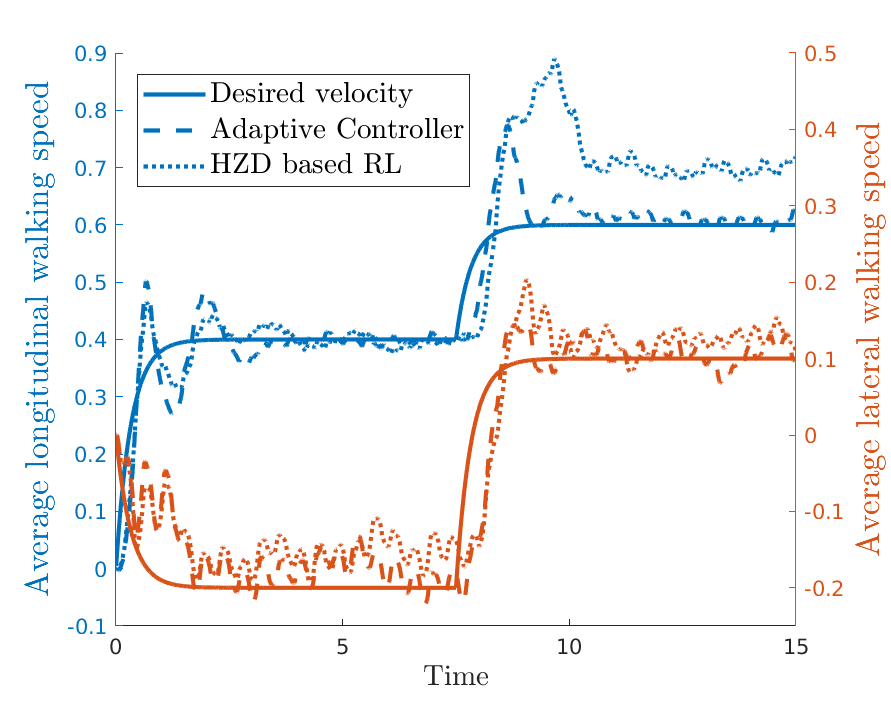}
\caption{Performance of the proposed adaptive controller compared with an HZD-based RL controller under changes in model properties when tracking desired velocities in both longitudinal and lateral directions at the same time.} 
\label{comparison_RLHZD_diag}
\vspace{-3mm}
\end{figure}

To further demonstrate the effectiveness of the adaptive controller to compensate for the unknown dynamics when tracking desired walking speeds in different directions we tested the controller for different walking speeds in diagonal directions. \figref{comparison_RLHZD_diag} shows the performance of the controller when tracking diagonal speed with $v_x = 0.4, v_y = -0.2$, and $v_x = 0.6, v_y = 0.1$,

% \begin{figure}
% \centering
% \vspace{1mm}
% \includegraphics[trim={0cm 0cm 0cm 0cm},clip,width=0.9\columnwidth]{figures/comparison_HZDRL_vy.eps}
% \caption{Velocity tracking performance in the lateral direction of the proposed adaptive controller compared against an HZD-based RL controller under the changes in model properties. The adaptive controller still shows a better tracking of desired velocities.} 
% \label{comparison_RLHZD_vy}
% % \vspace{-3mm}
% \end{figure}

% To test changes in different dynamic properties of the robot, we also modified the damping of the robot's joint by a factor of 10, which represents the case in which the joints of the robot wear out due to continuous use and generate more friction during its motion. As shown in \figref{damping}. the adaptive controller compensates the change in joints damping, demonstrating the capability of the controller to compensate for different types of model uncertainties.

\subsection{Robustness}

\begin{figure}
\centering
\vspace{1mm}
\includegraphics[trim={0cm 0cm 0cm 0cm},clip,width=0.9\columnwidth]{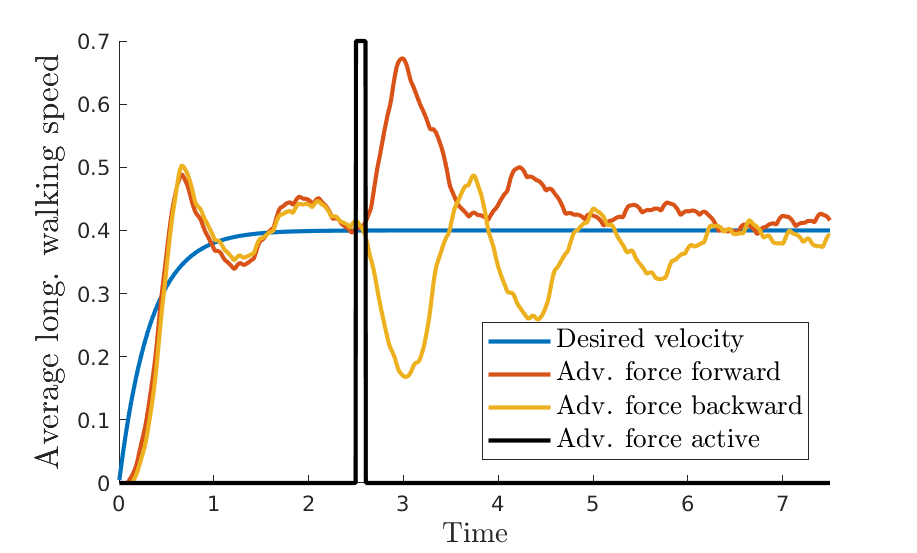}
\caption{The response of the walking motion using the proposed controller when various adversarial forces are applied in the forward and backward directions.} 
\label{adversarial}
\vspace{-3mm}
\end{figure}

Two tests were performed to evaluate the robustness of the adaptive controller, i) external disturbance rejections, and ii) walking on uneven terrain. To evaluate external disturbance rejection, an adversarial force is applied directly at the robot's torso in both forward and backward directions 2.5 seconds after the test started. \figref{adversarial}, shows the response of the adaptive controller when an adversarial force of $30N$ and a force of $25N$ are applied during 0.1 seconds in the forward and backward directions, respectively. The controller can handle both external forces successfully without falling and, more importantly, recovering the speed tracking performance quickly after the disturbance is applied.

\begin{figure}
\centering
\vspace{1mm}
\includegraphics[trim={0cm 0cm 0cm 0cm},clip,width=0.9\columnwidth]{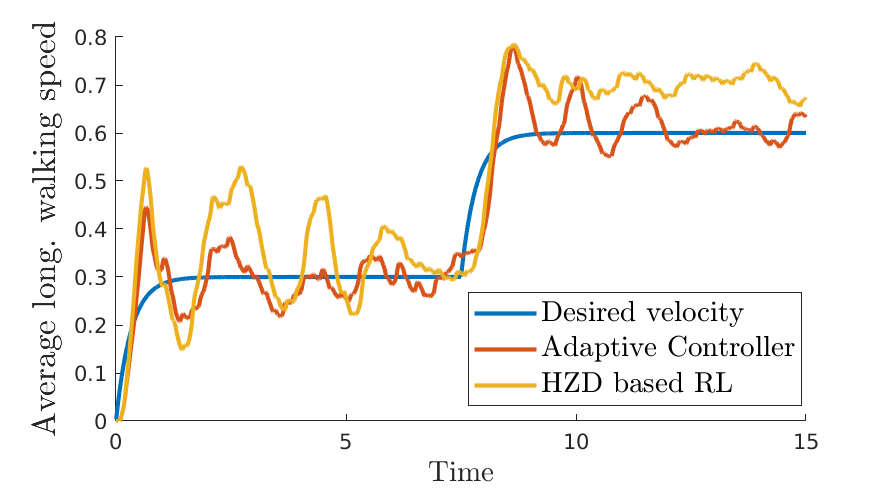}
\caption{Performance of the adaptive controller compared against a HZD-based RL controller when changing dynamic properties of the robot while walking on uneven terrain.} 
\label{terrain}
\vspace{-3mm}
\end{figure}

\figref{terrain} illustrates the response of the adaptive controller when the robot is walking on uneven terrain. The terrain presents significant irregularities with slopes up to $20$ degrees with a maximum height of $0.4 m$. The controller adapts successfully to the different terrain changes and keeps close tracking of the desired velocity throughout the whole test.  

% \begin{figure}
% \centering
% \vspace{1mm}
% \includegraphics[trim={0cm 0cm 0cm 0cm},clip,width=0.9\columnwidth]{figures/damping.eps}
% \caption{Performance of the adaptive controller compared against a HZD-based RL controller when the robot's joints damping is changed.} 
% \label{damping}
% \vspace{-3mm}
% \end{figure}

\subsection{Adaptation redundancy}

\begin{figure}
\centering
\vspace{1mm}
\includegraphics[trim={0cm 0cm 0cm 0cm},clip,width=0.9\columnwidth]{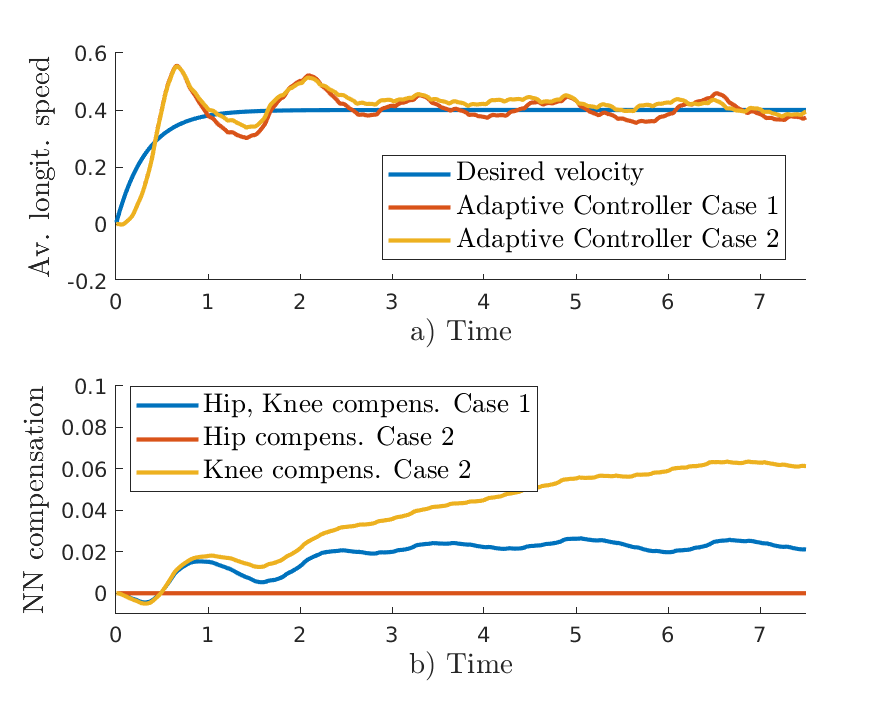}
\caption{Effect of neural network compensation to different joints in the adaptation capability of the proposed controller.} 
\label{adaptivetest}
\vspace{-3mm}
\end{figure}

In this subsection, we demonstrate the adaptability of the proposed adaptive controller framework to adapt to different operating conditions taking advantage of the generalized structure of the controller. In particular, \figref{adaptivetest} (a) shows the speed tracking performance of the adaptive controller for two cases: 1) both hip and knee joints use the compensation provided by the neural network to regulate the walking speed, and 2) the output of the neural network is forced to zero to test the adaptability of the controller to unknown scenarios. The results of this test are shown in \figref{adaptivetest} (b), where we see that for case 1, both stance hip and stance knee joints contribute equally to the compensation of the unknown dynamics while tracking the desired speed of the robot. For case 2, since we force the neural network output corresponding to the hip joint to zero, the adaptive controller learns a different way to compensate for the unknown dynamics of the system by only using the compensation available in the knee joint, which demonstrate the adaptation capabilities of the proposed controller.

\section{Conclusion}
\label{sec:conclusion}
This paper presents a general adaptive neural network-based controller for velocity tracking of 3D bipedal robots under model uncertainties. The proposed adaptive controller builds upon the concept of the virtual constraint of HZD-based controllers to incorporate adaptive trajectory compensation using neural networks to compensate for the unknown dynamics of the system. The result is a structure of simple yet effective adaptive neural network based controllers applied in a decoupled manner to render stable and robust limit walking cycles for effectively tracking desired walking velocity in both longitudinal and lateral directions. The adaptive controller shows the online learning process of the neural network is effective in compensating different changes in the dynamic properties of the system, such as the torso mass and the center of mass position of the torso. Moreover, the controller can learn different adaptation techniques such as using the knee joint instead of the hip joint for compensating the unknown dynamics. Improved performance of disturbance rejections---in the forms of adversarial forces and uneven terrains---are also observed with the proposed controller. The future work will focus on implementing the adaptive feedback controller on actual robots in experiments.

% \newpage
\bibliography{bib/IEEEabrv.bib, bib/ms.bib}
\bibliographystyle{IEEEtran}

\end{document}